\pdfoutput=1

\documentclass[sigconf]{acmart}
\renewcommand\footnotetextcopyrightpermission[1]{} 
\AtBeginDocument{%
  \providecommand\BibTeX{{%
    \normalfont B\kern-0.5em{\scshape i\kern-0.25em b}\kern-0.8em\TeX}}}

\setcopyright{acmcopyright}
\copyrightyear{2024}
\acmYear{2024}
\acmDOI{XXXXXXX.XXXXXXX}

\acmPrice{15.00}
\acmISBN{978-1-4503-XXXX-X/18/06}

\usepackage{multirow}
\usepackage{bbding}
\usepackage{bm}
\usepackage{enumitem}

\usepackage{colortbl}
\usepackage{color, xcolor}

\acmSubmissionID{123-A56-BU3}

\begin{document}

\title{Think-in-Memory: Recalling and Post-thinking Enable LLMs with Long-Term Memory}


\author{Lei Liu}
\authornote{Work was done when Lei Liu was a research intern at Ant Group.}
\email{liulei1497@gmail.com}
\orcid{0000-0001-8109-5248}
\affiliation{%
  \institution{CUHK-Shenzhen, Ant Group}
  \country{}
}

\author{Xiaoyan Yang}
\email{joyce.yxy@antgroup.com}
\affiliation{
  \institution{Ant Group}
  \country{}
}

\author{Yue Shen}
\authornote{Corresponding Author.}
\email{zhanying@antgroup.com}
\affiliation{
  \institution{Ant Group}
  \country{}
}

\author{Binbin Hu, Zhiqiang Zhang}
\email{{bin.hbb, lingyao.zzq}@antfin.com}
\affiliation{
  \institution{Ant Group}
  \country{}
}

\author{Jinjie Gu}
\email{jinjie.gjj@antfin.com}
\affiliation{
  \institution{Ant Group}
  \country{}
}

\author{Guannan Zhang}
\email{zgn138592@antfin.com}
\affiliation{
  \institution{Ant Group}
  \country{}
}



\begin{abstract}

Memory-augmented Large Language Models (LLMs) have demonstrated remarkable performance in long-term human-machine interactions, which basically relies on iterative recalling and reasoning of history to generate high-quality responses. However, such repeated recall-reason steps easily produce biased thoughts, \textit{i.e.}, inconsistent reasoning results when recalling the same history for different questions. On the contrary, humans can keep thoughts in the memory and recall them without repeated reasoning. Motivated by this human capability, we propose a novel memory mechanism called TiM (Think-in-Memory) that enables LLMs to maintain an evolved memory for storing historical thoughts along the conversation stream. The TiM framework consists of two crucial stages: (1) before generating a response, a LLM agent recalls relevant thoughts from memory, and (2) after generating a response, the LLM agent post-thinks and incorporates both historical and new thoughts to update the memory. Thus, TiM can eliminate the issue of repeated reasoning by saving the post-thinking thoughts as the history. Besides, we formulate the basic principles to organize the thoughts in memory based on the well-established operations, (\textit{i.e.}, insert, forget, and merge operations), allowing for dynamic updates and evolution of the thoughts. Furthermore, we introduce Locality-Sensitive Hashing into TiM to achieve efficient retrieval for the long-term conversations. We conduct qualitative and quantitative experiments on real-world and simulated dialogues covering a wide range of topics, demonstrating that equipping existing LLMs with TiM significantly enhances their performance in generating responses for long-term interactions.
\end{abstract}




\keywords{Large Language Model, Response Generation, Long-term Memory}



\maketitle

\section{Introduction}

Impressive advancements in Large Language Models (LLMs) have revolutionized the interaction between human beings and artificial intelligence (AI) systems. These advancements have particularly showcased superior performance in human-agent conversations, as demonstrated by ChatGPT \cite{chatgpt} and GPT-4 \cite{gpt-4}. From finance \cite{yang2023fingpt} and healthcare \cite{zhang2023huatuogpt} to business and customer service \cite{eloundou2023gpts}, these advanced LLMs exhibit a remarkable ability to understand questions and generate corresponding responses. Notably, the large model scale, reaching up to hundreds of billions of parameters, enables the emergence of such human-like abilities within LLMs \cite{kojima2022large}.


Despite the remarkable abilities of LLMs pre-trained on large corpora, LLM-based AI agents still face a significant limitation in long-term scenarios, \textit{i.e.}, inability to process exceptionally lengthy inputs \cite{liu2022relational}. This is particularly important in some specific tasks, \textit{e.g.}, medical AI assistants \cite{zhang2023huatuogpt} rely on the symptoms of past conversations to provide accurate clinical diagnosis. Thus, LLMs without the capability of dealing with long-term inputs may hinder the diagnosis accuracy due to forgetting important disease symptoms (see in Section \ref{application}). Therefore, it is necessary to develop AI systems with long-term capabilities for more accurate and reliable interactions.

\begin{figure*}[!t]
    \centering
    \includegraphics[width=1\linewidth]{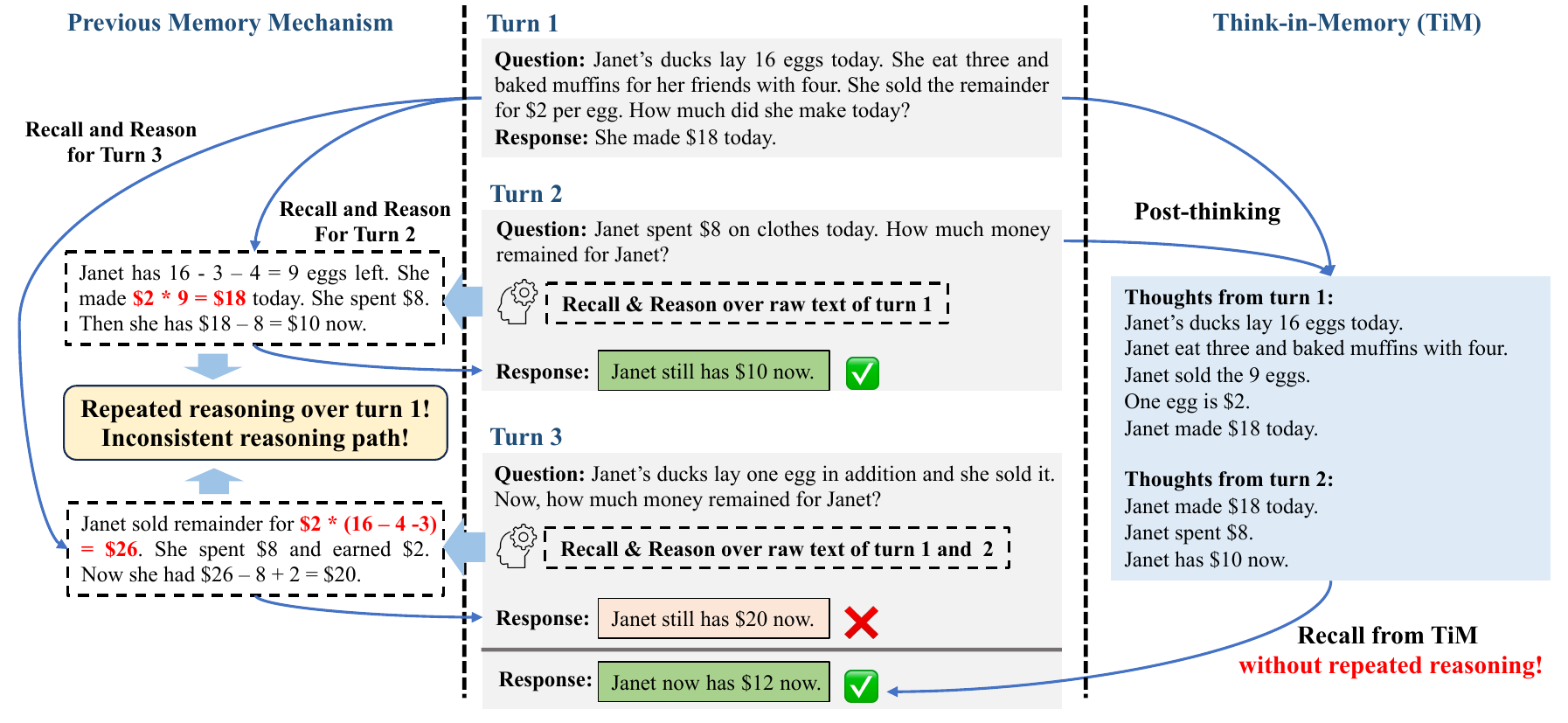}
    \caption{Comparisons between previous memory mechanisms with our proposed TiM. \textbf{(Left)}: Existing memory mechanisms mainly save raw text of previous turns, which require repeated reasoning over the same history. This easily leads to the inconsistent reasoning path ({\color{red} \textit{i.e.}, red part of the left}) with wrong response. \textbf{(Right)}: The proposed TiM stores the thoughts of LLMs for previous turns, which can avoid such inconsistency without repeated reasoning ({\color{red} \textit{i.e.}, red part of the right}).}
    \label{fig:issue}
\end{figure*}

There have been various studies conducted to improve the capabilities of LLMs to handle long-term inputs. Overall, these studies can be roughly divided into two types: (1) \textbf{Internal memory based methods} \cite{fournier2023practical} aims to reduce the computational costs of self-attention for expanding the sequence length. To accommodate longer input texts, special positional encoding should be exploited to learn relative positions. For example, \cite{phang2022investigating} explored a block-local Transformer with global encoder tokens, combined with additional long input pre-training. (2) \textbf{External memory based methods} (also called long-term memory mechanism \cite{wang2023augmenting}) generally utilize a physical space as a memory cache to store historical information, where relevant history can be read from the memory cache to augment LLMs without forgetting. In particular, both token and raw text can be maintained as history in the memory. For instance, \cite{borgeaud2022improving} demonstrated a significant performance improvement by augmenting LLMs with an external memory cache containing trillions of tokens assisted by BERT embeddings \cite{kenton2019bert}. It should be noticed that token-based memory mechanism requires to adjust the LLM's architecture for adaption, which is hard to be combined with different LLMs. By accessing an external memory cache, the augmented LLMs have achieved new state-of-the-art records in various language modeling benchmarks, which generally performs better than internal memory based methods. Therefore, this work focuses on designing an LLM-agnostic external memory mechanism to enhanced the memorization capacity of LLMs.



In general, the utility of memory-augmented LLMs primarily hinges on their ability for iterative recalling and repeated reasoning over the history in an external memory cache. In detail, for conversations after the $n$-th turn, LLMs are required to re-understand and re-reason the history from $0$-th to $(n-1)$-th conversations. For example, as shown in Figure \ref{fig:issue}, to answer the questions of $2$-th and $3$-th turns, LLMs recall $1$-th turn and reason over it for twice. Unfortunately, this paradigm is prone to encountering several issues and potentially causes a performance bottleneck in real-world applications. The main issues are shown in follows:

\begin{itemize}[leftmargin=*]
    \item \textbf{Inconsistent reasoning paths.} Prior studies \cite{adiwardana2020towards,wang2022self} has shown that LLMs easily generate diverse reasoning paths for the same query. As shown in Figure \ref{fig:issue} (Left), LLMs give a wrong response due to inconsistent reasoning over the context.
    \item \textbf{Unsatisfying retrieval cost.} To retrieve relevant history, previous memory mechanisms need to calculate pairwise similarity between the question and each historical conversation, which is time-consuming for long-term dialogue.
\end{itemize}

\begin{figure*}[!t]
    \centering
    \includegraphics[width=1\linewidth]{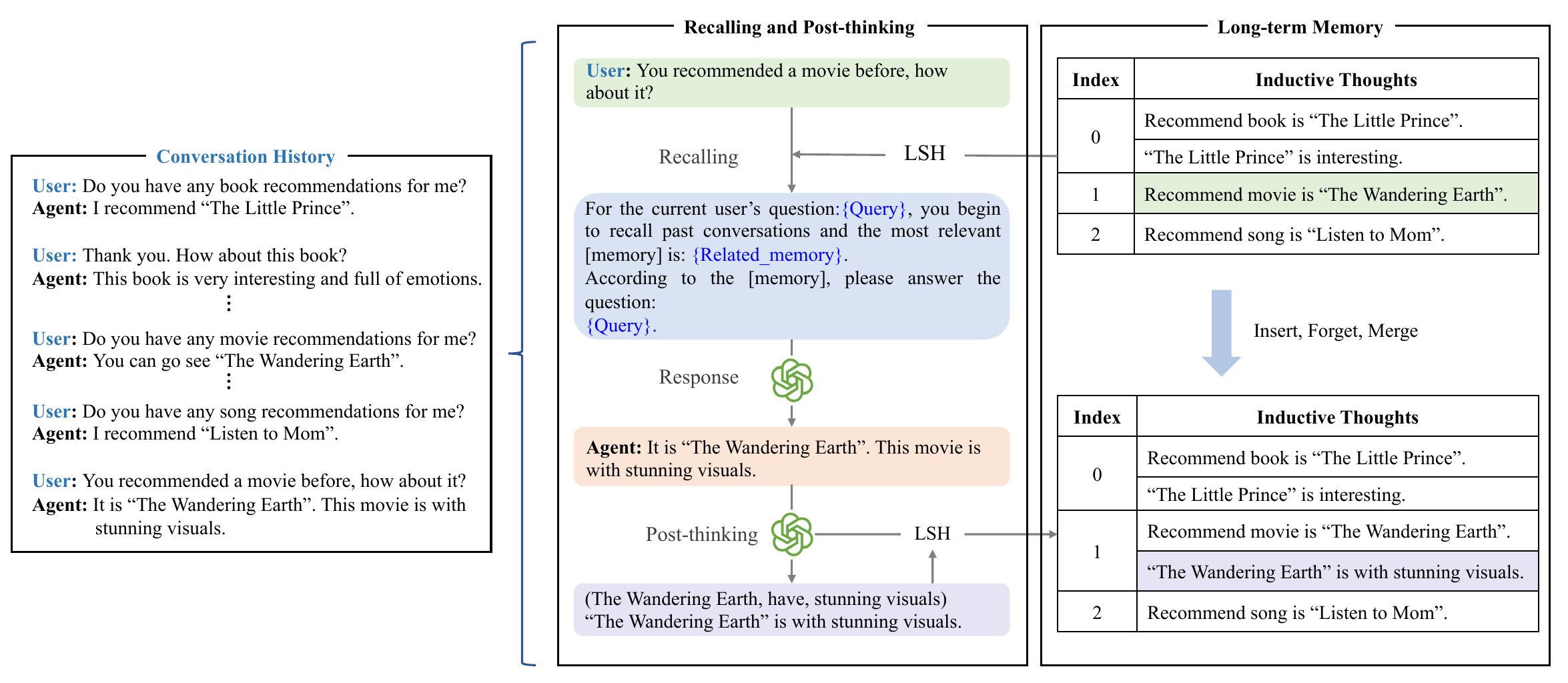}
    \caption{The overview of TiM framework. LLMs firstly recall history and give response for the question. Then new thoughts can be generated via the post-thinking step. These thoughts are saved as the memory to avoid repeated reasoning on the history.}
    \label{fig:framework}
\end{figure*}

To address these concerns, we would like to advance one step further in memory-augmented LLMs with the analogy to the typical process of metacognition~\cite{dunlosky2008metacognition}, where the brain saves thoughts as memories rather than the details of original events. Thus, in this work, we propose a \textbf{T}hink-\textbf{i}n-\textbf{M}emory (TiM) framework to model the human-like memory mechanism, which enables LLMs to remember and selectively recall historical thoughts in long-term interaction scenarios. Specifically, as shown in Figure \ref{fig:framework}, the TiM framework is divided into two stages: (1) In the recalling stage, LLMs generate the response for the new query with recalling relevant thoughts in the memory; (2) In the post-thinking stage, the LLM engages in reasoning and thinking over the response and saves new thoughts into an external memory. Besides, to mirror the cognitive process of humans, we formulate some basic principles to organize the thoughts in memory based on the well-established operations (\textit{e.g.}, insert, forget, and merge operations), allowing for dynamic updates and evolution of the thoughts. Specifically, TiM is built on a hash-based retrieval mechanism (\textit{i.e.}, Locality-Sensitive Hashing \cite{andoni2015practical}) to support efficient hand-in (\textit{i.e.}, insert thoughts) and hand-out (\textit{i.e.}, recall thoughts) operations. Additionally, TiM is designed to be LLM-agnostic, which means it can be combined with various types of language models. This includes closed-source LLMs such as ChatGPT \cite{chatgpt}, as well as open-source LLMs like ChatGLM\cite{zeng2022glm}.

The key contributions of this work are summarized as follows:
\begin{itemize}
    \item We propose a novel human-like long-term memory mechanism called TiM, enabling LLMs to remember and selectively recall thoughts. TiM can let LLM think in memory without repeated reasoning over the long-term history. 
    
    \item We formulate some basic principles to organize the thoughts in memory based on the well-established operations, which mirrors human cognitive process to empower dynamic updates and evolution for the thoughts in memory. Besides, a hash-based retrieval mechanism is introduced for efficient utilization of TiM.

    \item We conducted extensive experiments on multi-turn dialogue datasets. The results indicate that our method can substantially enhance LLM's performance across various dimensions: (1) It enables diverse topics ranging from open to specific domains; (2) It supports bilingual languages in both Chinese and English; (3) It improves response correctness and coherence.
\end{itemize}

\section{Related Work}


\subsection{Large Language Models}
Recently, Large Language Models (LLMs) have attracted significant attention for their superior performance on a wide range of Natural Language Processing tasks, such as machine translation \cite{zhang2023prompting}, sentiment analysis \cite{zhang2023enhancing}, and question answering systems \cite{guo2023images}. These advancements are indeed supported by the developments of deep learning techniques and the availability of vast amounts of text data. From the perspective of open source, existing LLMs can roughly divided into two types: (1) cutting-edge closed-source LLMs, \textit{e.g.}, PaLM \cite{chowdhery2022palm}, GPT-4 \cite{gpt-4}, and ChatGPT \cite{chatgpt}; (2) open-source LLMs, \textit{e.g.}, LLaMa \cite{touvron2023llama}, ChatGLM \cite{zeng2022glm}, and Alpaca \cite{taori2023stanford}. Researchers have studied various methods for the applications of these popular LLMs. For example, many strategies are proposed to fine-tune pre-trained LLM models on specific tasks \cite{min2023recent}, which can further improve their capabilities in specific domains. Besides, some efforts have been made to enhance the quality of the generated content of LLMs, \textit{e.g.}, generating more diverse and creative text while maintaining coherence and fluency \cite{dathathri2019plug}.
Overall, recent developments of LLMs cover a broad range of topics, including model architecture \cite{zeng2022glm}, training methods \cite{korbak2023pretraining}, fine-tuning strategies \cite{hu2021lora}, as well as ethical considerations \cite{chowdhery2022palm}. All these methods aim to enhance the understanding capabilities of LLMs for real-world applications. However, these powerful LLM models still have some shortcomings. One notable limitation of LLMs is their lack of a strong long-term memory, which hinders their ability to process lengthy context and retrieve relevant historical information.

\subsection{Long-term Memory}

Numerous efforts have been conducted to enhance the memory capabilities of LLMs. One approach is to utilize memory-augmented networks (MANNs) \cite{meng2018dialogue}, such as Neural Turing Machines (NTMs) \cite{graves2014neural}, which is designed to utilize more context information for dialogue. In general, MANNs are proposed with an external memory cache via the storage and manipulation of information, which can well handle tasks of long-term period by interacting with memory. In addition, many studies focused on long-term conversations \cite{xu2021beyond,xu2022long,zhong2023memorybank,liang2023unleashing}. For example, Xu \textit{et al.} \cite{xu2021beyond} introduced a new English dataset consisting of multi-session human-human crowdworker chats for long-term conversations. Zhong \textit{et al.} \cite{zhong2023memorybank} proposed a MemoryBank mechanism inspired by Ebbinghaus' forgetting curve theory. However, these methods still face some great challenges to achieve a reliable and adaptable long-term memory mechanism for Language and Learning Models (LLMs). Concretely, these methods only considered storing the raw dialogue text, requiring repeated reasoning of the LLM agent over the same history. Besides, these models need to calculate pairwise similarity for recalling relevant information, which is time-consuming for the long-term interactions.

\section{Methodology}
In this section, we first introduce the overall workflow of our proposed framework. Then we provide a detailed description for each stage of TiM, involving storage for memory cache, organization principle for memory updating, and retrieval for memory recalling.

\subsection{Framework Overview}
Given a sequence of conversation turns, each turn is denoted by a tuple $(Q, R)$, representing the user's query (Q) and the agent's response (R) at that specific turn. The main objective is to generate a more accurate response $R_y$ like a human for a new coming query $Q_x$, while remembering the contextual information of historical conversation turns. The proposed TiM allows the agent to process long-term conversation and retain useful historical information after multiple conversations with the user. 

\subsubsection{\textbf{Main Components}} As illustrated in Figure \ref{fig:framework}, our TiM comprises the following components, working together to provide more accurate and coherent responses for long-term conversation:
\begin{itemize}[leftmargin=*]
    \item Agent $\mathcal{A}$: $\mathcal{A}$ is a pre-trained LLM  model to facilitate dynamic conversations, such as ChatGPT \cite{chatgpt} and ChatGLM \cite{zeng2022glm}.
    \item Memory Cache $\mathcal{M}$: $\mathcal{M}$ a continually growing hash table of key-value pairs, where key is the hash index and value is a single thought. More details of $\mathcal{M}$ can refer to Section \ref{cache}. To be clear, $\mathcal{M}$ supports varying operations as shown in Table \ref{organ}. 
    \item Hash-based Mapping $\mathbb{\mathbf{F}}(\cdot)$: Locality-sensitive Hashing is introduced to quickly save and find the relevant thoughts in $\mathcal{M}$.
\end{itemize}

\subsubsection{\textbf{Workflow}} Overall framework is divided into two stages: 
\begin{itemize}[leftmargin=*]
    \item \textbf{Stage-1: Recall and Generation.} Given a new question from the user, LLM agent $\mathcal{A}$ retrieves relevant thoughts for generating accurate responses. Since we save the self-generated reasoning thoughts as external memory, this stage can directly recall and answer the question without repeated reasoning over the raw historical conversation text.
    \item \textbf{Stage-2: Post-think and Update.} After answering the question, we let the LLM agent post-think upon $Q$-$R$ pair and insert the newly self-generated reasoning thoughts into memory cache $\mathcal{M}$.
\end{itemize}

\subsection{Storage for Memory Cache}
\label{cache}
\subsubsection{Thoughts-based System}
TiM’s storage system $\mathcal{M}$ aims to save the knowledge of AI-user interactions via self-generated inductive thoughts (Definition \ref{thoughts}) upon the conversations. Each piece of thought $T$ is stored in the format of the tuple $(H_{idx}, T)$, where $H_{idx}$ is the hash index obtained by hash function $\mathbb{\mathbf{F}}(T)$. This hash-based storage not only aids in quick memory retrieval but also facilitates the memory updating, providing a detailed index of historical thoughts.

\begin{definition}
    \label{thoughts} \textbf{Inductive Thought.} The inductive thought is defined as the text which contains the relation between two entities, \textit{i.e.}, satisfying a relation triple $(E_h, r_i, E_t)$. $E_h$ is head entity connected with tail entity $E_t$ via the relation $r_i$, where $i \in [0, N]$ and $N$ is the relation number. Conceptually, $R_h=\{r_1, \cdots, r_N\}$ consists of all the one-hop relations for the entity $E_h$. 
\end{definition}

\begin{figure}[!t]
    \centering
    \includegraphics[width=1\linewidth]{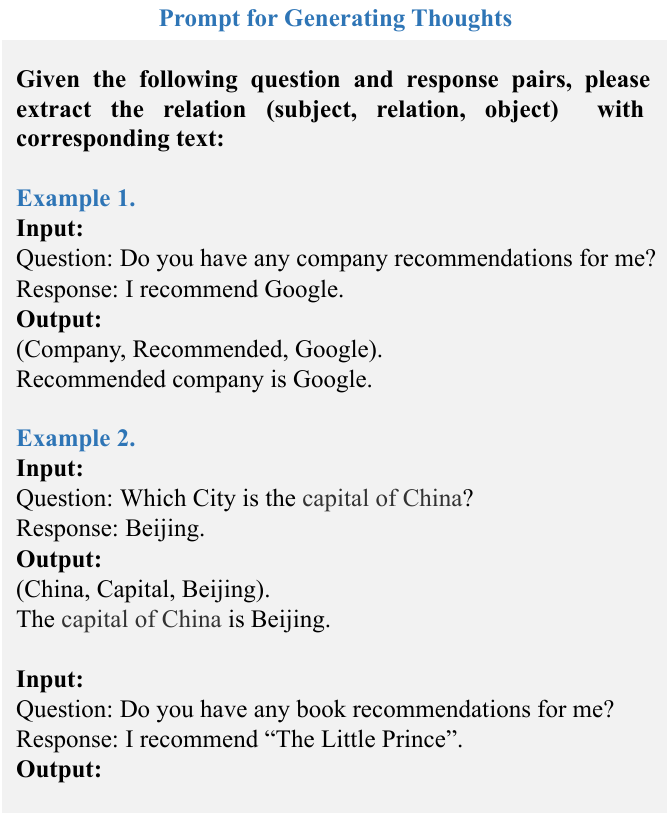}
    \caption{An example of prompts for generating thoughts.}
    \label{fig:triple_prompt}
\end{figure}

The main challenge of utilizing inductive thoughts for LLM is obtaining high-quality sentences matching relation triples. Here we provide two kinds of solutions to obtain inductive thoughts: (1) pre-trained model for open information extraction, such as OpenIE \cite{angeli2015leveraging}; (2) In-context learning with few-shot prompts based on LLM. In this work, we utilize the second solution, \textit{i.e.}, utilizing LLM Agent $\mathcal{A}$ to generate inductive thoughts, as shown in Figure \ref{fig:triple_prompt}.

\subsubsection{Hash-based Storage} We aim to save inductive thoughts into the memory following a certain rule, \textit{i.e.}, similar thoughts should be stored in the same group in the memory for efficiency. To this end, we adopt a hash table as the architecture of TiM’s storage system, where similar thoughts are assigned with the same hash index. 

Given a query, we propose to quickly search its nearest thoughts in a high-dimensional embedding space, which can be solved by the locality-sensitive hashing (LSH) method. The hashing scheme of LSH is to assign each $d$-dimension embedding vector $x\in \mathbf{R}^d$ to a hash index $\mathbf{F}(x)$, where nearby vectors get the same hash index with higher probability. We achieve this by exploiting a random projection as follows:
\begin{equation}
    \mathbb{\mathbf{F}}(x) = \mathop{\arg\max}\left(\left[xR; -xR\right]\right),
    \label{LSH_eq}
\end{equation}
where $R$ is a random matrix of size $(d, b/2)$ and $b$ is the number of groups in the memory. $\left[u; v\right]$ denotes the concatenation of two vectors. This LSH method is a well known LSH scheme \cite{andoni2015practical} and is easy to implement. Figure \ref{fig:framework} shows a schematic exhibition of TiM’s storage system based on LSH.

\begin{figure}[!t]
    \centering
    \includegraphics[width=0.9\linewidth]{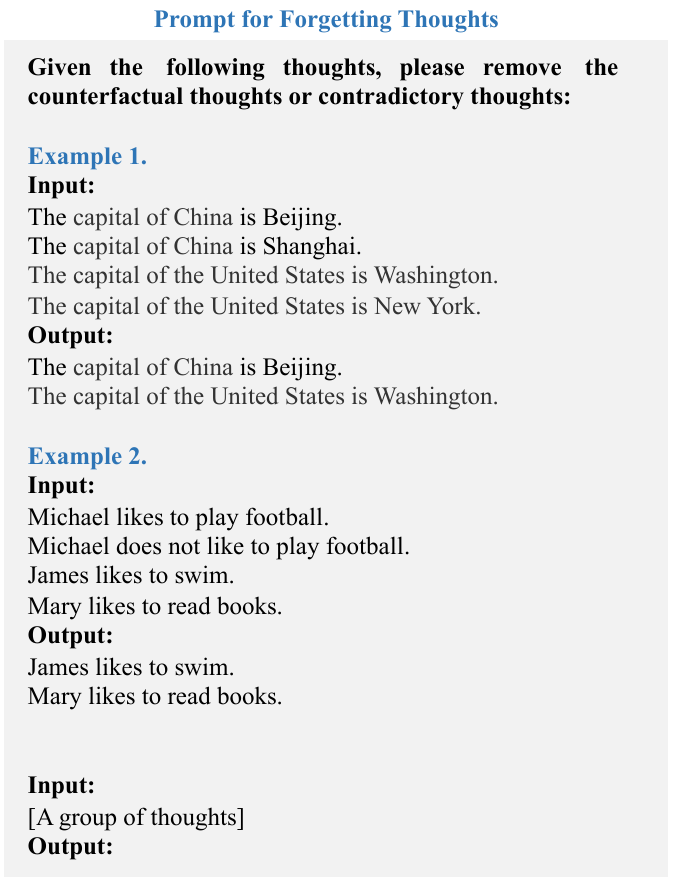}
    \caption{An example of prompts for forgetting thoughts.}
    \label{fig:forget_prompt}
\end{figure}

\begin{figure}[!t]
    \centering
    \includegraphics[width=0.9\linewidth]{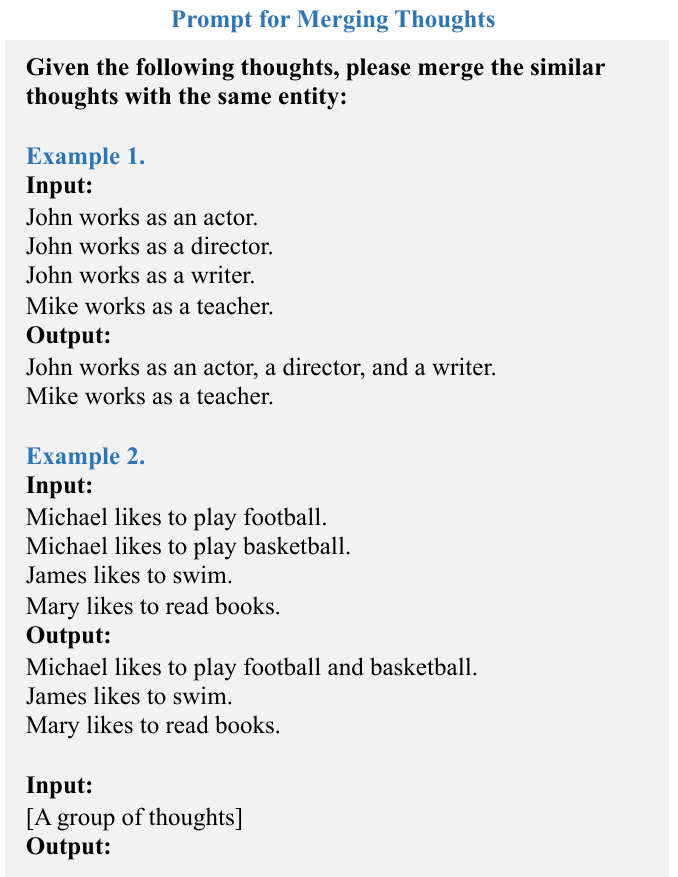}
    \caption{An example of prompts for merging thoughts.}
    \label{fig:merge_prompt}
\end{figure}

\subsection{Retrieval for Memory Recalling}
Built on the memory storage, the memory retrieval operates a two-stage retrieval task for the most relevant thoughts, \textit{i.e.}, LSH-based retrieval followed by similarity-based retrieval. The paradigm involves the following detailed points.

\begin{itemize}[leftmargin=*]
    \item \textbf{Stage-1: LSH-based Retrieval.} For a new query $Q$, we first obtain its embedding vector $x$ based on LLM agent. Then LSH function (\textit{i.e.}, Eq. \ref{LSH_eq}) can produce the hash index of the query. This hash index also indicates the its nearest group for similar thoughts in the memory cache according to the property of LSH.
    \item \textbf{Stage-2: Similarity-based Retrieval.} Within the nearest group, we calculate the pairwise similarity between the query and each piece of thought in the group. Then top-$k$ thoughts are recalled as the relevant history for accurately answering the query. It should be noticed that pairwise similarity is calculated within a group rather than the whole memory cache, which can achieve more efficient retrieval than previous memory mechanisms.
    
\end{itemize}

\subsection{Organization for Memory Updating}
With the above-discussed memory storage and retrieval, the long-term memory capability of LLMs can be well enhanced. Motivated by the human memory, there needs some organization principles based on the well-established operations for dynamic updates and evolution of the thoughts, \textit{e.g.}, insert new thoughts, forget less important thoughts, and merge repeated thoughts, which can make the memory mechanism more natural and applicable. 

Beginning with the architecture of the storage for memory cache, TiM adopts the hash table to store the self-generated thoughts, where each hash index corresponds a group containing similar thoughts. Within same group, TiM supports the following operations to organize the thoughts in the memory:
\begin{itemize}[leftmargin=*]
    \item \textbf{Insert}, \textit{i.e.}, store new thoughts into the memory. The prompt for generating thoughts is shown in Figure \ref{fig:triple_prompt}.

    \item \textbf{Forget}, \textit{i.e.}, remove unnecessary thoughts from the memory, such as contradictory thoughts. The prompt of this operation is shown in Figure \ref{fig:forget_prompt}.

    \item \textbf{Merge}, \textit{i.e.}, merge similar thoughts in the memory, such as thoughts with the same head entity. The prompt of this operation is shown in Figure \ref{fig:merge_prompt}.
\end{itemize}

\subsection{Parameter-efficient Tuning}

We adopt a computation-efficient fine-tuning approach called Low-Rank Adaptation (LoRA) \cite{hu2021lora} for the scenarios with limited computational resources. LoRA \cite{hu2021lora} optimizes pairs of rank-decomposition matrices while keeping the original weights frozen, which can effectively reduce the number of trainable parameters. Specifically, considering a linear layer defined as $y = Wx$, LoRA fine-tunes it according to $y = Wx + BAx$, where $W \in \mathbf{R}^{d\times k}$, $B \in \mathbf{R}^{d\times r}$, $A \in \mathbf{R}^{r\times k}$, and $r \ll \min(d; k)$. Essentially, this fine-tuning stage can adapt LLMs to multi-turn conversations, providing appropriately and effectively response to users. For all experiments, we set LoRA rank $r$ as 16 and train the LLM models for $10$ epochs.

\begin{table}[!t]
  \caption{Organization comparisons between previous memory mechanisms and ours. KG denotes the knowledge graph and Q-R denotes the question and response pairs.}
  \label{organ}
  \resizebox{\linewidth}{!}{
  \begin{tabular}{lccccc}
    \toprule
    Method & Content & LLM-agnostic & Insert & Forget & Merge \\
    \midrule
    SCM \cite{liang2023unleashing}       & Q-R        & \Checkmark    & \Checkmark & \XSolidBrush &  \XSolidBrush            \\
    RelationLM \cite{liu2022relational} & KG  & \XSolidBrush  & \Checkmark & \XSolidBrush &  \XSolidBrush            \\
    LongMem \cite{wang2023augmenting}    & Token           & \XSolidBrush  & \Checkmark & \XSolidBrush &  \XSolidBrush            \\
    MemoryBank \cite{zhong2023memorybank} & Q-R        & \Checkmark    & \Checkmark & \Checkmark   &  \XSolidBrush            \\
    \midrule
    Ours (TiM) & Thoughts         & \Checkmark    & \Checkmark & \Checkmark   &  \Checkmark              \\
  \bottomrule
\end{tabular}}
\end{table}

\subsection{Insightful Discussion}
Here we make a summary for previous memory mechanisms and our method in Table \ref{organ}, including memory content, LLM-agnostic, and organization operations. There are several important observations from Table \ref{organ}: (1) Previous memory mechanisms only save raw conversation text (Q-R pairs) as the memory, which requires repeated reasoning over the history. Our method maintains thoughts in the memory cache and can directly recall them without repeated reasoning. (2) Previous memory mechanisms only support simple read and write (insert) operations, while our method provides more manipulate way for the memory. (3) Some previous memory mechanisms store the tokens in the memory, which requires adjusting LLM architecture (LLM-aware) for applications. Our method is deigned as a LLM-agnostic module, which can be easily combined with other LLMs.

\begin{table*}[!t]
    \caption{Comparison Results on Three Datasets. Top-5 thoughts are recalled from the memory cache.}
    \label{main_result}
    \begin{tabular}{c|c|c|c|c|c|c}
    \toprule
    Dataset              & LLM                       & Language/Topic                & Memory             & Retrieval Accuracy & Response Correctness & Contextual Coherence    \\ \midrule
    \multirow{4}{*}{GVD} & \multirow{4}{*}{ChatGLM}  & \multirow{2}{*}{English/Open} & SiliconFriend      & 0.809              & 0.438                & 0.680           \\ \cmidrule{4-7}
                         &                           &                               & TiM (Ours)         & \textbf{0.820}     & \textbf{0.450}       & \textbf{0.735}  \\ \cmidrule{3-7}
                         &                           & \multirow{2}{*}{Chinese/Open} & SiliconFriend      & 0.840              & 0.418                & 0.428           \\ \cmidrule{4-7}
                         &                           &                               & TiM (Ours)         & \textbf{0.850}     & \textbf{0.605}       & \textbf{0.665}  \\ \midrule \midrule             
    \multirow{12}{*}{Kdconv} & \multirow{6}{*}{ChatGLM}  & \multirow{2}{*}{Chinese/Film} & \XSolidBrush       & -                  & 0.657                & 0.923   \\ \cmidrule{4-7}
                         &                           &                         & TiM (Ours)         & \textbf{0.920}     & \textbf{0.827}       & \textbf{0.943}   \\ \cmidrule{3-7}
                         &                           & \multirow{2}{*}{Chinese/Music}& \XSolidBrush       & -                  & 0.666                & 0.910    \\ \cmidrule{4-7}
                         &                           &                         & TiM (Ours)         & \textbf{0.970}     & \textbf{0.826}       & \textbf{0.926}   \\ \cmidrule{3-7}  
                         &                           & \multirow{2}{*}{Chinese/Travel}& \XSolidBrush      & -                  & 0.735                & 0.906    \\ \cmidrule{4-7}
                         &                           &                         & TiM (Ours)         & \textbf{0.940}     & \textbf{0.766}       & \textbf{0.912}   \\ \cmidrule{2-7}
                         & \multirow{6}{*}{Baichuan2}& \multirow{2}{*}{Chinese/Film} & \XSolidBrush       & -                  & 0.360                & 0.413   \\ \cmidrule{4-7}
                         &                           &                         & TiM (Ours)         & \textbf{0.913}     & \textbf{0.743}       & \textbf{0.870}   \\ \cmidrule{3-7}
                         &                           & \multirow{2}{*}{Chinese/Music}& \XSolidBrush       & -                  & 0.253                & 0.283    \\ \cmidrule{4-7}
                         &                           &                         & TiM (Ours)         & \textbf{0.900}     & \textbf{0.710}       & \textbf{0.780}   \\ \cmidrule{3-7}  
                         &                           & \multirow{2}{*}{Chinese/Travel}& \XSolidBrush      & -                  & 0.207                & 0.280    \\ \cmidrule{4-7}
                         &                           &                         & TiM (Ours)         & \textbf{0.833}     & \textbf{0.757}       & \textbf{0.807}   \\ \midrule \midrule 
  
    \multirow{4}{*}{RMD} & \multirow{2}{*}{ChatGLM}  & \multirow{2}{*}{Chinese/Medical}& \XSolidBrush      & -                  & 0.806                & 0.893      \\\cmidrule{4-7}
                         &                                 &                          & TiM (Ours)         & \textbf{0.900}      & \textbf{0.843}       & \textbf{0.943}  \\ \cmidrule{2-7}
                         & \multirow{2}{*}{Baichuan2}  & \multirow{2}{*}{Chinese/Medical}& \XSolidBrush      & -                  & 0.506                & 0.538      \\\cmidrule{4-7}
                         &                                 &                          & TiM (Ours)         & \textbf{0.873}      & \textbf{0.538}       & \textbf{0.663}  \\ \midrule
                         
    \end{tabular}
\end{table*}

\section{Experiment}
\subsection{Experimental Settings}
\subsubsection{Dataset}
Three datasets are used to demonstrate the effectiveness of the proposed method.
\begin{itemize}[leftmargin=*]
    \item \textbf{KdConv}: KdConv is a Chinese multi-domain knowledge-driven conversation benchmark \cite{zhou-etal-2020-kdconv} grounding the topics to knowledge graphs, which involves 4.5K conversations and 86K utterances from three domains (film, music, and travel). The average turn number is 19.

    \item \textbf{Generated Virtual Dataset (GVD)}: GVD is a long-term conversation dataset \cite{zhong2023memorybank} involving 15 virtual users (ChatGPT) over 10 days. Conversations are synthesized using pre-defined topics, including both English and Chinese languages. For the test set, \cite{zhong2023memorybank} manually constructed 194 query questions (97 in English and 97 in Chinese) to evaluate whether the LLM could accurately recall the memory and produce the appropriate answers.

    \item \textbf{Real-world Medical Dataset (RMD)}: To evaluate the effectiveness of the proposed memory mechanism in the real-world scenarios, we manually collect and construct a dataset containing 1,800 conversations for medical healthcare consumer. For the test set, 80 conversations are used to evaluate whether the LLM could provide the accurate diagnosis.
    
\end{itemize}

\subsubsection{LLM}
We integrate two powerful LLMs to demonstrate the effectiveness of the proposed TiM mechanism. These LLMs originally lack long-term memory and specific adaptability to the long-term conversations. The detailed introduction of these LLMs are follows.
\begin{itemize}[leftmargin=*]
    \item \textbf{ChatGLM} \cite{zeng2022glm}: ChatGLM is an open-source bilingual language model based on the General Language Model (GLM) framework \cite{zeng2022glm}. This model contains 6.2 billion parameters with specific optimization, involves supervised fine-tuning, feedback bootstrap, and reinforcement learning with human feedback. 
    
    \item \textbf{Baichuan2} \cite{yang2023baichuan}: Baichuan2 is an open-source large-scale multilingual language model containing 13 billion parameters, which is trained from scratch on 2.6 trillion tokens. This model excels at dialogue and context understanding.
\end{itemize}

\subsubsection{Baseline}
One baseline is to answer questions without using any memory mechanism. Another baseline is SiliconFriend \cite{zhong2023memorybank}, a classical memory mechanism, which can store the raw text into the memory and support reading operation.

\subsubsection{Evaluation Protocol}
Following \cite{zhong2023memorybank}, three metrics are adopted to evaluate the performance of the proposed method.
\begin{itemize}[leftmargin=*]
    \item \textbf{Retrieval Accuracy} evaluates whether the relevant memory is successfully recalled (labels: \{0: no; 1: yes\}).
    
    \item \textbf{Response Correctness} evaluates if correctly answering the probing question (labels: \{0: wrong; 0.5: partial; 1 : correct\}).

    \item \textbf{Contextual Coherence} evaluates whether the response is naturally and coherently generated, \textit{e.g.}, connecting the dialogue context and retrieved memory (labels: \{0: not coherent; 0.5: partially coherent; 1: coherent\}).
\end{itemize}
To be fair, during evaluation, the prediction results of all LLMs are firstly shuffled, ensuring the human evaluator does not know which LLM the results come from. Then the final evaluation results are obtained by the human evaluation. 

\subsection{Comparison Results}

\subsubsection{Results on GVD dataset} 
We evaluate our method on both English and Chinese test sets of GVD dataset. The following insights are observed from Table \ref{main_result}: (1) Compared with SiliconFriend \cite{zhong2023memorybank}, our method exhibits superior performance for all metric, especially for the contextual coherence, indicating the effectiveness of TiM mechanism. (2) TiM delivers better results on both languages. The performance improvement on Chinese is larger than English, which may be attributed to the abilities of the LLMs.

\subsubsection{Results on KdConv dataset} Table \ref{main_result} illustrates the comparison results on KdConv dataset. We evaluate 2 different LLMs with TiM over different topics (film, music, and travel). As shown in Table \ref{main_result}, it is observed that our method can obtain best results across all topics. Our method can achieve high retrieval accuracy to recall the relevant thoughts. When without the memory mechanism, these LLMs usually exhibit lower response correctness due to lack of long-term memory capability, while TiM can well eliminate such negative issue. Furthermore, TiM can also help to improve the contextual coherence of the response.

\subsubsection{Results on RMD dataset} Table \ref{main_result} reports the comparison results on RMD dataset, which contains the realistic conversations between the doctors and patients. As shown in Table \ref{main_result}, our method can improve the overall response performance for the real-world medical conversations. In detail, using TiM, both ChatGLM and Baichuan2 can improve their capability for long-term conversations, \textit{i.e.}, significant improvements on the response correctness and the contextual coherence. The main reason is that TiM is more similar to the workflow of human memory, which can enhance the ability of LLMs to produce more human-like responses. 

\begin{table}[!t]
  \caption{Comparisons of Retrieval Time. Baseline calculates pairwise similarity between the question and memory.}
  \label{time}
  \begin{tabular}{c|c}
    \toprule
    Method & Retrieval Time (ms)  \\
    \midrule
    Baseline   &   0.6287         \\ \midrule
    Ours (TiM) &   0.5305         \\
  \bottomrule
\end{tabular}
\end{table}

\begin{figure}[!t]
    \setlength{\abovecaptionskip}{0cm}
    \centering
    \includegraphics[width=1\linewidth]{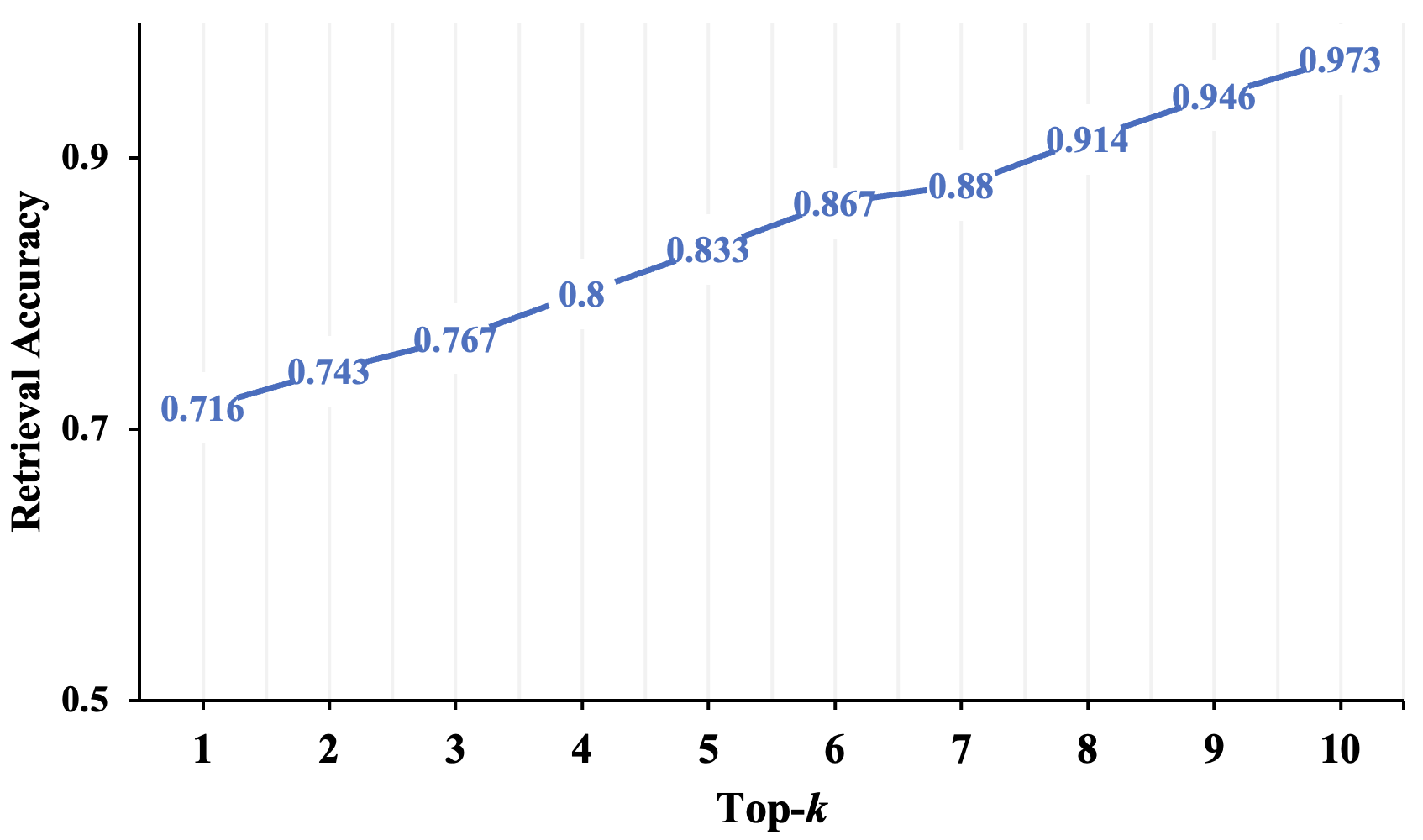}
    \caption{Tendency of retrieval accuracy with different $k$.}
    \label{fig:topk}
\end{figure}

\begin{figure*}[!t]
    \centering
    \setlength{\abovecaptionskip}{0cm}
    \includegraphics[width=1\linewidth]{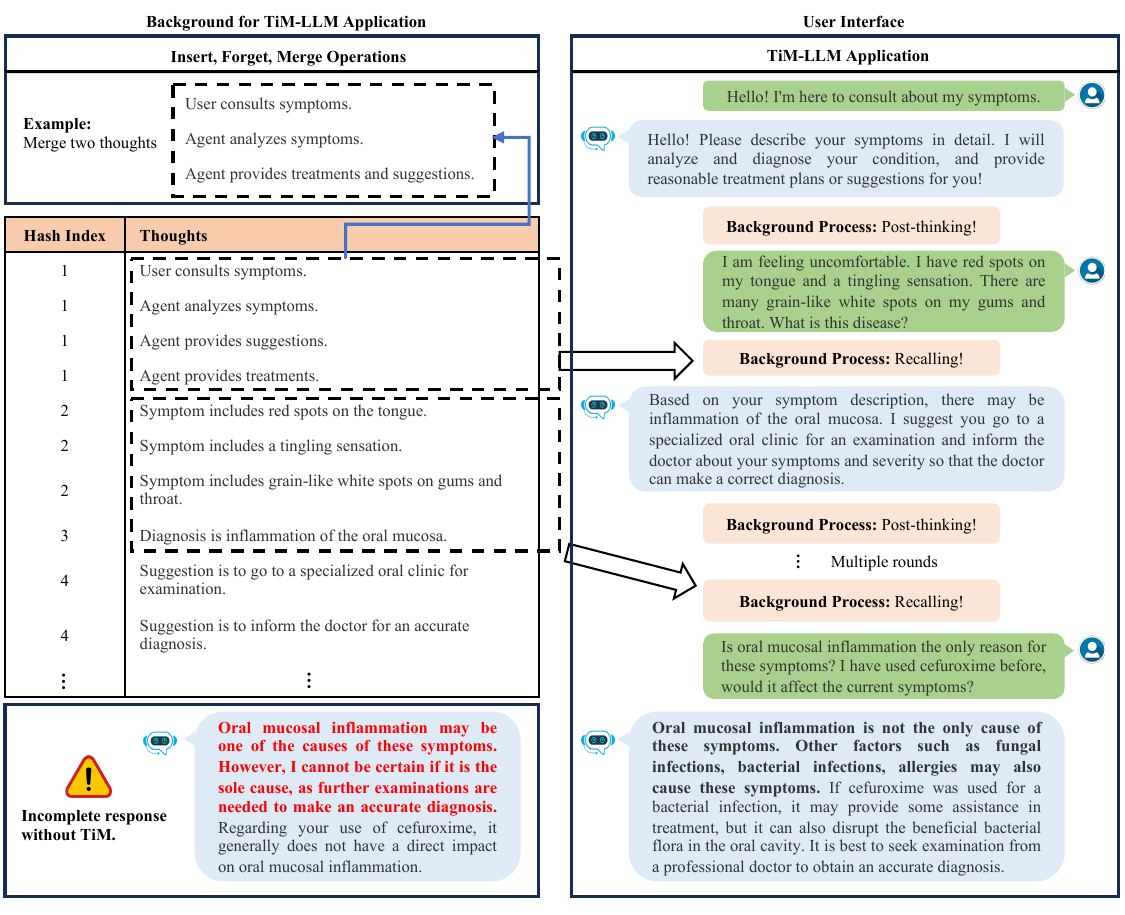}
    \caption{The application of TiM. The left is the background of TiM-LLM application and the right is user interface.}
    \label{fig:app}
\end{figure*}

\subsection{More Analysis}
\subsubsection{Retrieval Time} We report the comparison results of retrieval time. The baseline is to calculate pairwise similarity between the question and the whole memory, which is utilized as the default retrieval way for most previous mechanisms. For both baseline and our method, the memory length is as $140$ and the memory context is fixed. Table \ref{main_result} shows the time cost for making a single retrieval. It is observed that our method can reduce about 0.1 ms retrieval time compared with baseline method.

\subsubsection{Top-$k$ Recall} We report the retrieval accuracy using different values of $k$ on Kdconv dataset (Travel). As shown in Figure \ref{fig:topk}, top-1 retrieval accuracy is higher than $0.7$. The overall retrieval accuracy is improved with increasing value of $k$, where top-$10$ can achieve $0.973$ retrieval accuracy. Besides, as shown in Table \ref{main_result}, top-$5$ recall can significantly improve the performance of existing LLMs for long-term conversations.

\subsection{Industry Application}
\label{application}
In this section, based on the ChatGLM and TiM, we develop a medical agent (named TiM-LLM) in the context of patient-doctor conversations (as shown in Figure \ref{fig:app}). Note that TiM-LLM is only an \textbf{auxiliary} tool for the clinical doctors to give treatment options and medical suggestions for patients' needs. 

Figure \ref{fig:app} illustrates a real-world conversation between a patient and a doctor, where the clinical diagnosis results are given by the medical agent with and without TiM. As shown in Figure \ref{fig:app}, without TiM, the medical agent may struggle to recall previous symptoms, resulting in incomplete or incorrect assessments ({\color{red}red part}), \textit{i.e.}, the agent has forgotten previous symptoms so it is uncertain whether oral mucosal inflammation is the only cause. Assisted by TiM, the medical agent can recall relevant symptoms and make a comprehensive understanding of a patient's diseases. Thus it provide accurate diagnosis and treatment (\textbf{bold part}). 


\section{Conclusion}
In this work, we propose a novel memory mechanism called TiM to address the issue of biased thoughts in Memory-augmented LLMs. By storing historical thoughts in an evolved memory, TiM enables LLMs to recall relevant thoughts and incorporate them into the conversations without repeated reasoning. TiM consists of two key stages: recalling thoughts before generation and post-thinking after generation. Besides, TiM works with the several basic principles to organize the thoughts in memory, which can achieve dynamic updates of the memory. Furthermore, we introduce Locality-Sensitive Hashing into TiM to achieve efficient retrieval for the long-term conversations. The qualitative and quantitative experiments conducted on real-world and simulated dialogues demonstrate the significant benefits of equipping LLMs with TiM. Overall, TiM is designed as an approach to improve the quality and consistency of responses for long-term human-AI interactions.

\newpage

\bibliographystyle{unsrt} 
\bibliography{sample-base}

\end{document}